\documentclass[conference,a4paper]{IEEEtran}
\usepackage{cite}
\usepackage{amsmath,amssymb,amsfonts}
\usepackage{algorithmic}
\usepackage{graphicx}
\usepackage{subfigure}
\usepackage{textcomp}
\usepackage{xcolor}
\def\BibTeX{{\rm B\kern-.05em{\sc i\kern-.025em b}\kern-.08em
    T\kern-.1667em\lower.7ex\hbox{E}\kern-.125emX}}
\begin{document}

\title{GUSOT: Green and Unsupervised Single Object Tracking for Long Video Sequences}

\author{\IEEEauthorblockN{
Zhiruo Zhou \IEEEauthorrefmark{1}, 
Hongyu Fu \IEEEauthorrefmark{1}, 
Suya You \IEEEauthorrefmark{2} and
C.-C. Jay Kuo \IEEEauthorrefmark{1}}
\IEEEauthorblockA{\IEEEauthorrefmark{1}University of Southern California,
Los Angeles, USA \\
}
\IEEEauthorblockA{\IEEEauthorrefmark{2}Army Research Laboratory, Maryland, USA \\
}
}

\maketitle

\begin{abstract}
Supervised and unsupervised deep trackers that rely on deep learning
technologies are popular in recent years.  Yet, they demand high
computational complexity and a high memory cost. A green unsupervised
single-object tracker, called GUSOT, that aims at object tracking for
long videos under a resource-constrained environment is proposed in this
work.  Built upon a baseline tracker, UHP-SOT++, which works well for
short-term tracking, GUSOT contains two additional new modules: 1)
lost object recovery, and 2) color-saliency-based shape proposal. They help resolve
the tracking loss problem and offer a more flexible object proposal,
respectively. Thus, they enable GUSOT to achieve higher tracking
accuracy in the long run. We conduct experiments on the large-scale
dataset LaSOT with long video sequences, and show that GUSOT offers a
lightweight high-performance tracking solution that finds applications
in mobile and edge computing platforms. 
\end{abstract}


\section{Introduction}\label{sec:introduction}

Video object tracking is a fundamental problem in computer vision and
has a wide range of applications such as autonomous navigation and video
recognition. A popular branch of visual tracking is single object
tracking where the object marked in the first frame is tracked through
the whole video. Deep-learning-based trackers, called deep trackers, have
been popular in the last 7 years.  Supervised deep trackers have been
intensively studied. Its superior performance is achieved by exploiting
a large amount of offline labeled data. While
supervision is powerful in guiding the learning process, it casts doubt
on the reliability of tracking unseen objects. Unsupervised deep trackers
\cite{wang2021unsupervisedcvpr, wu2021progressive, zheng2021learning,
shen2022unsupervised, zhou2021uhp, zhou2021unsupervised} have been
developed to address this concern in recent years. 

Research on supervised and unsupervised deep trackers has primarily
focused on tracking performance. The high performance of deep trackers
is accompanied with high computational complexity and a huge memory
cost. Generally, they are difficult to deploy in resource-limited
platforms such as mobile and edge devices. Specific examples include
drones, autonomous vehicles, mobile phones, etc.  Furthermore, some
state-of-the-art trackers are short-term trackers since they cannot
recover from object tracking loss automatically, which occurs frequently
in long video tracking scenarios. 

To tackle the unsupervised long-term tracking problem in a
resource-constrained environment, we propose a green and unsupervised
single-object tracker (GUSOT) in this work.  Built upon a short-term
tracking baseline known as UHP-SOT++ \cite{zhou2021unsupervised}, we
introduce two additional new modules to GUSOT: 1) lost object recovery,
and 2) color-saliency-based shape proposal.  The first module helps
recover a lost object with a set of candidates by leveraging motion in
the scene and selecting the best one with local/global features (e.g.,
color information). The second module facilitates accurate and long-term
tracking by proposing bounding-box proposals of flexible shape for the
underlying object using low-cost yet effective segmentation. Both
modules are lightweight and can be easily integrated with UHP-SOT++.
They enable GUSOT to achieve higher tracking accuracy in the long run.
We conduct experiments on a large-scale benchmark dataset, LaSOT
\cite{fan2019lasot}, containing long video sequences and compare GUSOT
with several state-of-the-art trackers.  Experimental results show that
GUSOT offers a lightweight tracking solution whose performance is
comparable with that of deep trackers. 

The rest of this paper is organized as follows. Related work is reviewed
in Sec. \ref{sec:review}. The proposed GUSOT method is detailed in Sec.
\ref{sec:method}. Experimental results are given in Sec.
\ref{sec:experiments}.  Finally, concluding remarks are provided in Sec.
\ref{sec:conclusion}. 

\begin{figure*}[htbp]
\centerline{\includegraphics[width=\textwidth]{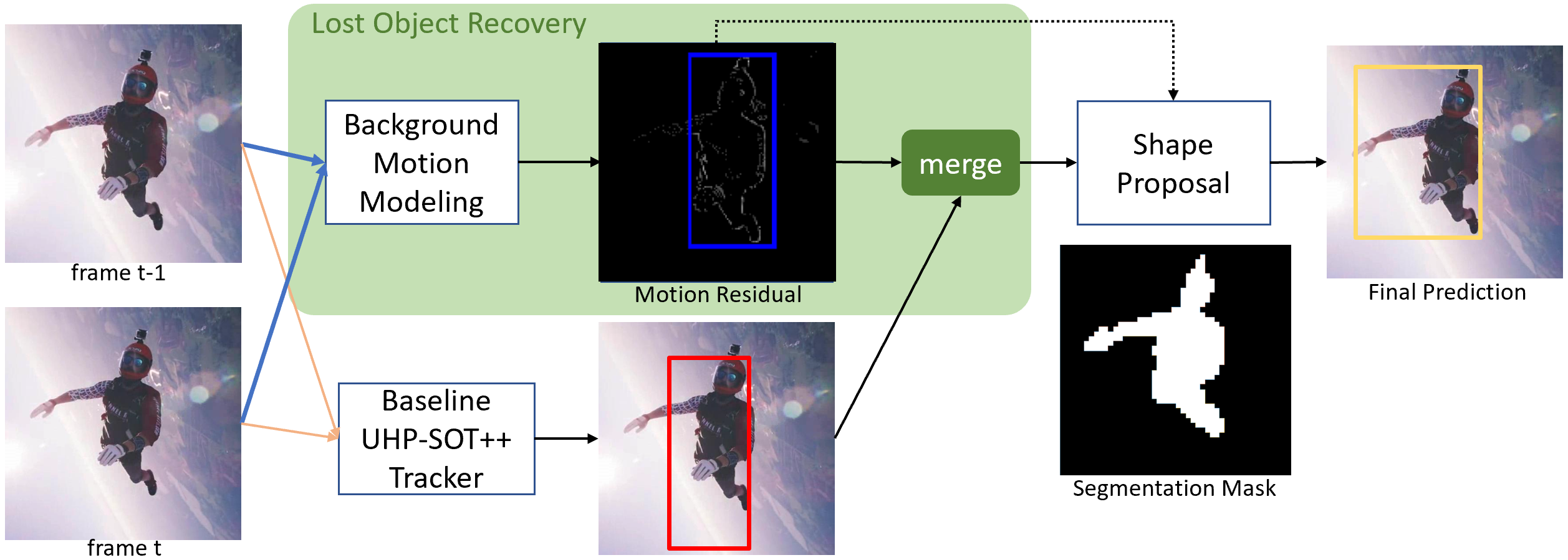}}
\caption{An overview of the proposed GUSOT tracker, where the red and
blue boxes denote the baseline and the motion proposals, respectively.
The one with higher appearance similarity is chosen to be the location
proposal. Then, the third proposal, called the shape proposal, is used
to adjust the shape of the location proposal.  The final predicted box
is depicted by the yellow box.}\label{fig:framework}
\end{figure*}

\section{Related Work}\label{sec:review}

{\bf Supervised deep trackers.} Supervised deep trackers
\cite{danelljan2017eco, danelljan2016beyond, ma2015hierarchical,
qi2016hedged, bhat2018unveiling, wang2018multi, sun2019roi, lu2018deep,
nam2016learning, pu2018deep, song2017crest} has progressed a lot in the
last seven years and offer state-of-the-art tracking accuracy.
Siamese-network-based trackers \cite{bertinetto2016fully, li2018high,
li2019siamrpn++, tao2016siamese, zhu2018distractor} are popular because
of their simple architecture yet effective tracking performance.  They
extract features from the target object as well as a search window
around the target using a shared deep network and, then, conduct
cross-correlation between the two feature maps to generate a response
map. The response map can be used to locate the object and further
processed by some downstream heads for object classification and box
regression.  Recently, supervised transformer-based trackers have been
developed in \cite{wang2021transformer, chen2021transformer} with
excellent performance. 

{\bf Unsupervised deep trackers.} Research on unsupervised deep trackers
focuses on training deep trackers without labels on offline datasets
\cite{wang2021unsupervisedcvpr, wu2021progressive, zheng2021learning,
shen2022unsupervised}. UDT \cite{wang2021unsupervisedcvpr} offers a
pioneering solution along this direction. It trains a network by
tracking forward and backward in video with cycle consistency. Later,
efforts have been made to improve the learning capability of UDT.
Examples include mining positive and negative samples in ResPUL
\cite{wu2021progressive}, mining moving objects via optical flow in USOT
\cite{zheng2021learning}, or improving cycle training with better
long-short term features or loss re-weighting in ULAST
\cite{shen2022unsupervised}.  Yet, many of them only have a limited
performance gain as compared with conventional trackers based on
discriminative correlation filters (DCFs).  Furthermore, they are not
lightweight trackers and cannot be adopted in resource-constrained
devices. 

{\bf Unsupervised conventional trackers.} Before the deep learning era,
most conventional trackers are unsupervised. One class of them are based
on DCFs \cite{bolme2010visual, henriques2014high,
danelljan2015convolutional, danelljan2016discriminative,
danelljan2016beyond, bertinetto2016staple, li2018learning,
xu2019learning, li2020autotrack}.  DCF trackers learn a template,
$\mathbf{f}$, from the first frame with regression and update the
template via
\begin{equation}\label{eq:regression}
\mathrm{arg}\min_{\mathbf{f}} \frac{1}{2}\|\sum_{d=1}^D \mathbf{x}^{d}*
\mathbf{f}^{d}-\mathbf{y}\|^2,
\end{equation}
where $\mathbf{x}$ is the representation of the object patch with $D$
features, $\mathbf{y}$ is the pre-defined regression label, and $*$ is
the feature-wise spatial convolution. The correlation between the
learned template and the search region is implemented by the fast
Fourier transform (FFT). It can be executed efficiently even on CPUs.
Handcrafted features such as the histogram of oriented gradients (HOG)
and colornames \cite{danelljan2014adaptive} are used for feature
extraction.  Yet, the performance of DCF trackers is inferior to that of
supervised deep trackers by a significant margin. 

{\bf UHP-SOT and UHP-SOT++.} UHP-SOT was proposed in \cite{zhou2021uhp}
to boost the performance of DCF trackers.  It adopted a DCF tracker,
STRCF \cite{li2018learning}, as the baseline and updated the template in
each frame by
\begin{eqnarray}
\mathrm{arg}\min_{\mathbf{f}} & \Big\{ & \frac{1}{2}\|\sum_{d=1}^D \mathbf{x}_{t}^{d}*
\mathbf{f}^{d}-\mathbf{y}\|^2 +  \frac{1}{2}\sum_{d=1}^D \|\mathbf{w}\cdot 
\mathbf{f}^{d} \|^2 \nonumber \\
&& + \frac{\mu}{2}\|\mathbf{f}-\mathbf{f}_{t-1} \|^2 \Big\}, \label{eq:strcf}
\end{eqnarray}
where $\mathbf{w}$ is the spatial weight on the template to suppress
background, $\mathbf{f}_{t-1}$ is the template learned from time $t-1$,
and $\mu$ is a constant regularization coefficient. With the STRCF
baseline, two modules (i.e., background motion modeling and
trajectory-based box prediction) were added for performance improvement.
UHP-SOT achieved a significant performance gain over STRCF on OTB-2015
\cite{7001050}. Its enhanced version, UHP-SOT++
\cite{zhou2021unsupervised}, has been tested on more datasets with
further performance improvement. We adopt UHP-SOT++ as the baseline in
GUSOT as elaborated in the next section. 

\section{Proposed GUSOT Method}\label{sec:method}

An overview of the proposed GUSOT method is given in
Fig.~\ref{fig:framework}. GUSOT adds two new modules to the baseline
UHP-SOT++ tracker for higher performance in long video tracking: 1) lost
object recovery and 2) color-saliency-based shape proposal. While the
baseline offers a box proposal (the red box), the recovery module yields
a motion residual map and provides the second box proposal of the same
size, called the motion proposal (the blue box).  The two proposals are
compared and the one with higher appearance similarity with a trusted
template is chosen as the location proposal.  Finally, another proposal,
called the shape proposal, is used to adjust the shape of the location
proposal to yield the final prediction (the yellow box).  For the
UHP-SOT++ baseline, we refer to \cite{zhou2021unsupervised}.  The
operations of the two new modules are detailed below. 

{\bf Lost Object Recovery.} When the baseline tracker works properly,
the lost object recovery module simply serves as a backup. Yet, its role
becomes critical when the baseline tracker fails.  Recall that most
trackers rely on similarity matching between the learned template
$\mathbf{f}$ at frame $(t-1)$ and the search region in frame $t$ to
locate the object.  However, if the object gets lost due to tracking
error or deformations, it is difficult to capture it again as the search
region often drifts away from the true object location.  Exhaustive
search over the whole frame is infeasible in online tracking.  Besides,
the learned template could be contaminated during the object loss period
and it cannot be used to search the object anymore. This module is used
to find object locations of higher likelihood efficiently and robustly. 

\begin{figure}[thbp]
\centering
\subfigure[random sampling]{\includegraphics[width=0.48\linewidth]{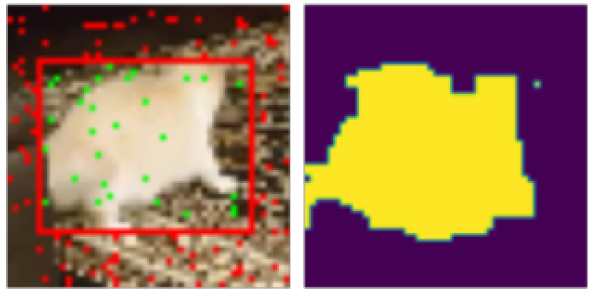}}
\subfigure[color-saliency-based sampling]{\includegraphics[width=0.48\linewidth]{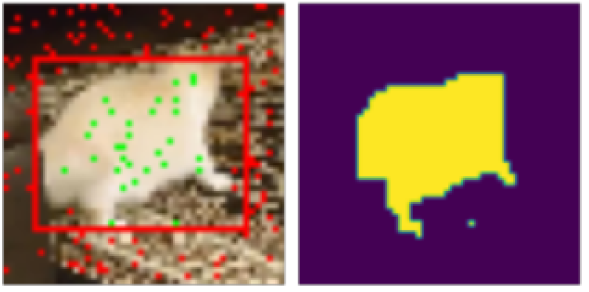}}
\caption{Comparison of two sampling scheme and their segmentation
results, where green and red dots represent foreground and background
initial points, respectively. The red box is the reference box which
indicates the object location.}\label{fig:sampling}
\end{figure}

\begin{figure}[thbp]
\centerline{\includegraphics[width=\linewidth]{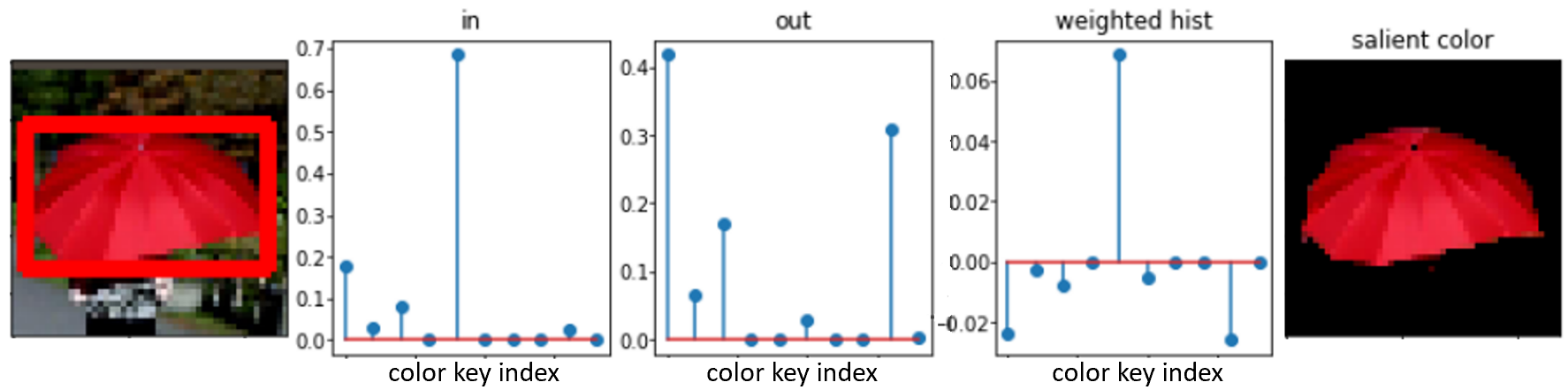}}
\caption{Determination of salient color keys.}\label{fig:colorkey}
\end{figure}

This module is built upon background motion estimation and compensation
\cite{zhou2021uhp}.  Based on the correspondence of sparsely sampled
background salient points between frame $(t-1)$ and frame $t$, we can
estimate the global motion field of the scene. Next, we can apply the
motion field to all pixels in frame $(t-1)$, which is called the motion
compensated frame, and find the difference between frame $t$ and
motion-compensated frame $(t-1)$, leading to a motion residual map.
Afterwards, a box proposal, which is of the same size as that of frame
$(t-1)$ and covering the largest amount of motion residuals, is
computed. It serves as a good proposal for the object since the object
can be revealed by residuals if 1) the object and its background take
different motion paths and 2) background motion is compensated. 

Now, we have two bounding box proposals: 1) the baseline proposal from
UHP-SOT++ and 2) the motion proposal as discussed above. We need to assess
their quality and select a better one. This is achieved by similarity
measure. To avoid template degradation from tracking loss, we store a
trusted template $\mathbf{f^*}$. It could be the initial template or a
learned template from a high-confidence frame.  Let $\mathbf{x}$ be the
candidate proposal. Then, the similarity measure between $\mathbf{f^*}$
and $\mathbf{x}$ can be defined as their correlation coefficient:
\begin{eqnarray}\label{eq:sim1}
s_{1}(\mathbf{f^*}, \mathbf{x}) = \frac{\langle \mathbf{f^*}, 
\mathbf{x} \rangle}{\|\mathbf{f^*}\| \|\mathbf{x}\|},
\end{eqnarray}
where $\langle \cdot , \cdot \rangle$ is the vector inner product,
$\mathbf{f^*}$ and $\mathbf{x}$ are feature representations (rather than
pixel values) of a trusted template or a candidate.  Commonly used
features are the histogram of oriented gradients (HOG) and colornames.
Here, we calculate a normalized histogram $v$ of color keys in
colornames. Then, the similarity of two histograms is measured using the
Chi-square distance:
\begin{eqnarray}\label{eq:sim2}
s_{2}(\mathbf{f^*}, \mathbf{x}) = \sum_{i}{\frac{(v_{\mathbf{f^*},i} - 
v_{\mathbf{x},i})^2}{v_{\mathbf{f^*},i} + v_{\mathbf{x},i}}}.
\end{eqnarray}
The two histograms are more similar if the Chi-square distance is smaller.
We replace the baseline proposal, $\mathbf{x_2}$, with the motion
proposal, $\mathbf{x_1}$, if 
$$
s_{1}(\mathbf{f^*},\mathbf{x_1}) > s_{1}(\mathbf{f^*}, \mathbf{x_2}) \mbox{ and } 
s_{2}(\mathbf{f^*}, \mathbf{x_1}) \leq s_{2}(\mathbf{f^*}, \mathbf{x_2}).
$$ 
Otherwise, we keep the baseline proposal.  For a faster tracking speed,
the lost object recovery module is turned on only when the similarity
score of the baseline proposal is low or the predicted object centers
show abnormal trajectories (e.g., getting stuck to a corner). 

{\bf Color-Saliency-Based Shape Proposal.} Good shape estimation can
benefit the tracking performance in the long run as it leads to tight
bounding boxes and thus reduces noise in template learning. Deep
trackers use the region proposal network to offer flexible boxes. Here,
we exploit several low-cost segmentation techniques for box shape
estimation in online tracking. 

Given an image of size $W\times H$, the foreground/background
segmentation is to assign binary labels $l_p=l(x,y) \in \{0,1\}$ to
pixels $p=(x,y)$. The binary mask, $\mathbf{I}$, can be estimated using
the Markov Random Field (MRF) optimization framework:
\begin{eqnarray}\label{eq:mrf}
\mathbf{I^*} = \arg \min_{\mathbf{I}}{\sum_{p}{\rho (p,l_p)} + 
\sum_{\{p,q\}\in \aleph}{w_{pq} \|l_p-l_q\|}},
\end{eqnarray}
where $\rho (p,l_p)$ is the cost of assigning $l_p$ to pixel $p$, which
is defined as the negative log-likelihood of $p$ in the Gaussian
mixtures, $\aleph$ is the four-connected neighborhood, $w_{pq}$ is the
weight of mis-matching penalty which is calculated as the Euclidean
distance between $p$ and $q$.  The first term of Eq. (\ref{eq:mrf})
ensures that similar pixels get the same label while its second term
forces label continuity among neighbors. The fast algorithm implemented
in \cite{chen2014fast, zehuamrfbcd} can work on a $32\times32$ image in
50ms. Thus, we crop a small patch centered at the predicted location and
conduct coarse segmentation.  

The MRF optimization problem is solved by iteration, which is to be
initialized by a set of foreground/background points of good quality.
Random sampling inside/outside the box results in noisy initialization
as shown in Fig.~\ref{fig:sampling}(a). It tends to lead to undesired
errors and/or longer iterations. To address it, we exploit salient
foreground and background colors.  With the predicted box and its
associated image patch, all colors on the patch are quantized into $N$
color keys. The distributions of color keys in and out of the object
box, denoted by $p_{in}$ and $p_{out}$, are calculated. Then, the
color saliency score (CSS) of color keys $k_i$, $i=1,\cdots, N$, is 
the weighted difference between the two distributions:
\begin{eqnarray}\label{eq:salient}
\mbox{CSS}(k_i) = \frac{\sum_{j\neq i}{\exp \|k_i-k_j\|^2}}{Z} (p_{in}(k_i)-p_{out}(k_i)).
\end{eqnarray}
$Z$ is the normalization factor for the sum of exponentials. The weights take the color difference into consideration and favor color keys standing out among all colors. If $\mbox{CSS}(k_i)$ is a positive (or negative) number of large
magnitude, $k_i$ is a foreground (or background) color key. A visualization example is provided in Fig.~\ref{fig:colorkey}. The
histogram bin number, $N$, is adaptively determined based on the first
frame. It first tries the color keys used in colornames features and
takes the corresponding number if there exists clear salient colors in
this setting. Otherwise, clustering of all colors and merging of
clusters are conducted to determine $N$. Finally, we sample points
according to their color saliency score. As shown in Fig.~\ref{fig:sampling},  color-saliency-based sampling provides better initial points for iteration, thus leading
to a better segmentation mask. 

\begin{figure}[htbp]
\centerline{\includegraphics[width=0.9\linewidth]{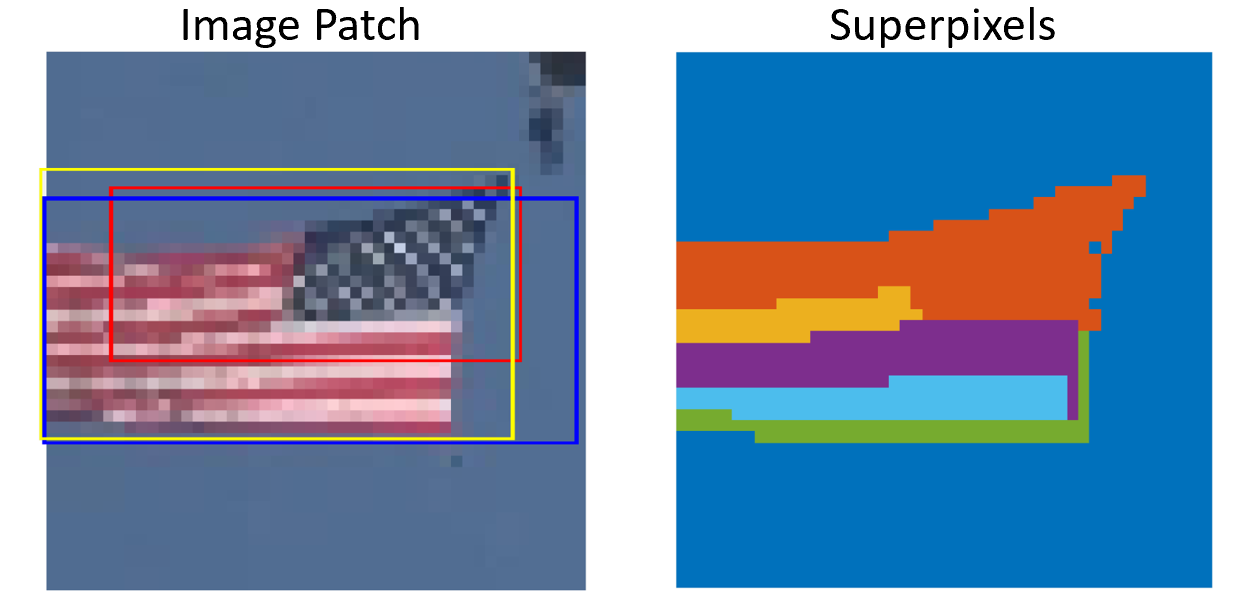}}
\caption{Illustration of shape proposal derivation based on superpixel
segmentation, where the red, blue and yellow boxes correspond to the
baseline, motion, and shape proposals, respectively. The size of the
motion proposal here is determined by clipping integral curves
horizontally and vertically on the motion residual map.} 
\label{fig:seg_sp}
\end{figure}

Complicated objects may not have clear salient colors. If the MRF
optimization generates an abnormal output, we switch to a simple
yet effective shape proposal by merging superpixels with guidance from
motion and baseline. As shown in Fig.~\ref{fig:seg_sp}, after superpixel
segmentation \cite{felzenszwalb2004efficient}, we assign binary labels
to superpixels and find the enclosing box, $B_s$, for the foreground
superpixels. With different label assignments, different $B_s$ can be generated.

{\bf Output Proposal.} Let $B_b$, $B_m$ and $B_s$ denote the baseline
proposal, the motion proposal and the shape proposal,
respectively.  Usually, we stick to baseline proposal $B_{b}$ if its
similarity score is high. We will only consider $B_{m}$, $B_{s}$, and
$B^*$ if the baseline gets a low similarity score.  When this happens,
we compute the following proposal as the predicted tracking box at frame
$t$:
\begin{eqnarray}\label{eq:box}
B^* = {\rm arg} \max_{B_{s}} {\mathrm{IoU}(B_{s},B_{b}) + \mathrm{IoU}(B_{s},B_{m})},
\end{eqnarray}
where IoU is the intersection-over-union between two boxes. 

\section{Experiments}\label{sec:experiments}

{\bf Experimental Setup.} We conduct experiments on the large-scale
single object tracking dataset LaSOT \cite{fan2019lasot}. It has 280
long test videos of around 685K frames. Evaluation metrics for tracking
performance include: 1) the distance precision (DP) measured at the
20-pixel threshold and 2) the area-under-curve (AUC) score for the
overlap precision. We use the same hyper-parameter settings for the
baseline tracker. The segmentation module is activated when the
appearance score of the baseline tracker is less than $0.2$.  Almost all
template matching trackers can provide the appearance score for an
object proposal. It is simply the correlation score in DCF-trackers.
The patch size used in segmentation is $48\times48$. The superpixel
segmentation exploits a Gaussian blur of $\sigma_{blur}=0.6$ and the
minimal superpixel size is set to 50. 

\begin{table}[htbp]
\caption{Comparison of unsupervised and supervised trackers on LaSOT,
where S and P indicate \textbf{S}upervised and \textbf{P}re-trained, 
respectively. Backbone denotes the pre-trained feature extraction network. }
\label{tab1:comp}
\begin{center}
\begin{tabular}{ccc|cc|cc}
\hline
 & S & P & DP & AUC & GPU & Backbone \\ \hline
ECO-HC\cite{danelljan2017eco}  & $\times$ & $\times$ & 27.9 & 30.4 & $\times$ & N/A \\
STRCF\cite{li2018learning}  & $\times$ & $\times$ & 29.8 & 30.8 & $\times$ & N/A \\
UHP-SOT++\cite{zhou2021unsupervised}  & $\times$ & $\times$ & 32.9 & 32.9 & $\times$ & N/A \\ 
\hline
LUDT\cite{wang2021unsupervised}  & $\times$ & \checkmark & - & 26.2 & \checkmark & VGG\cite{chatfield2014return} \\
USOT\cite{zheng2021learning}  & $\times$ & \checkmark & 32.3 & 33.7 & \checkmark & ResNet50\cite{he2016deep} \\
ULAST\cite{shen2022unsupervised}  & $\times$ & \checkmark & 40.7 & 43.3 & \checkmark & ResNet50\cite{he2016deep} \\ 
\hline
SiamFC\cite{bertinetto2016fully}  & \checkmark & \checkmark & 33.9 & 33.6 & \checkmark & AlexNet\cite{krizhevsky2012imagenet} \\
ECO\cite{danelljan2017eco}  & \checkmark & \checkmark & 30.1 & 32.4 & \checkmark & VGG\cite{chatfield2014return} \\
SiamRPN\cite{li2018high}  & \checkmark & \checkmark & 38.0 & 41.1 & \checkmark & AlexNet\cite{krizhevsky2012imagenet} \\ 
\hline\hline
GUSOT (Ours)  & $\times$ & $\times$ & 36.1 & 36.8 & $\times$ & N/A \\ \hline
\end{tabular}
\end{center}
\end{table}

\begin{figure*}[htbp]
\centerline{\includegraphics[width=\textwidth]{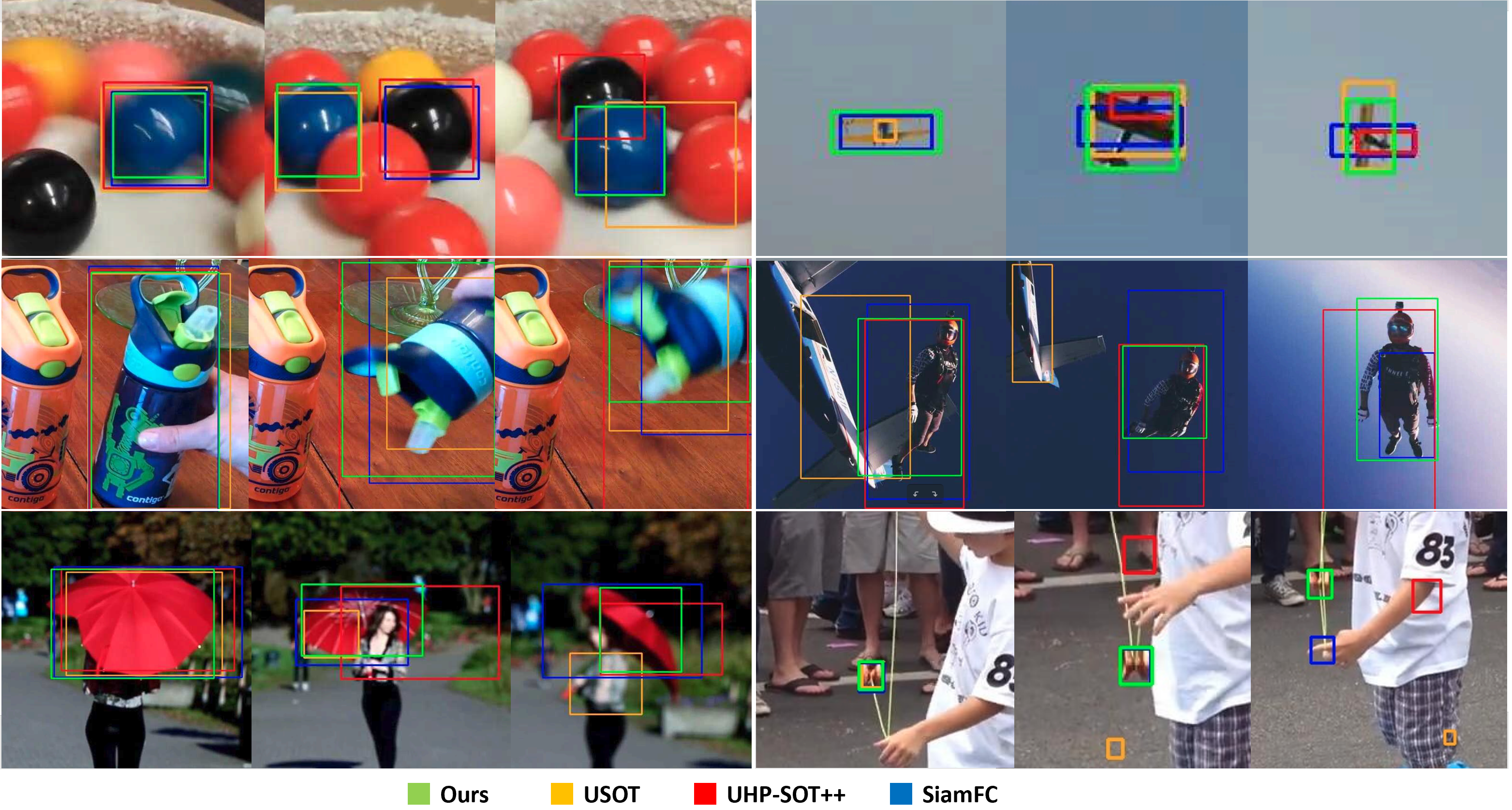}}
\caption{Qualitative evaluation of GUSOT, UHP-SOT++, USOT and SiamFC.
From left to right and top to bottom, the sequences presented are
\textit{pool-12}, \textit{airplane-15}, \textit{bottle-14},
\textit{person-10}, \textit{umbrella-2} and \textit{yoyo-17},
respectively.} \label{fig:vis}
\end{figure*}

\begin{figure}[htbp]
\centerline{\includegraphics[width=\linewidth]{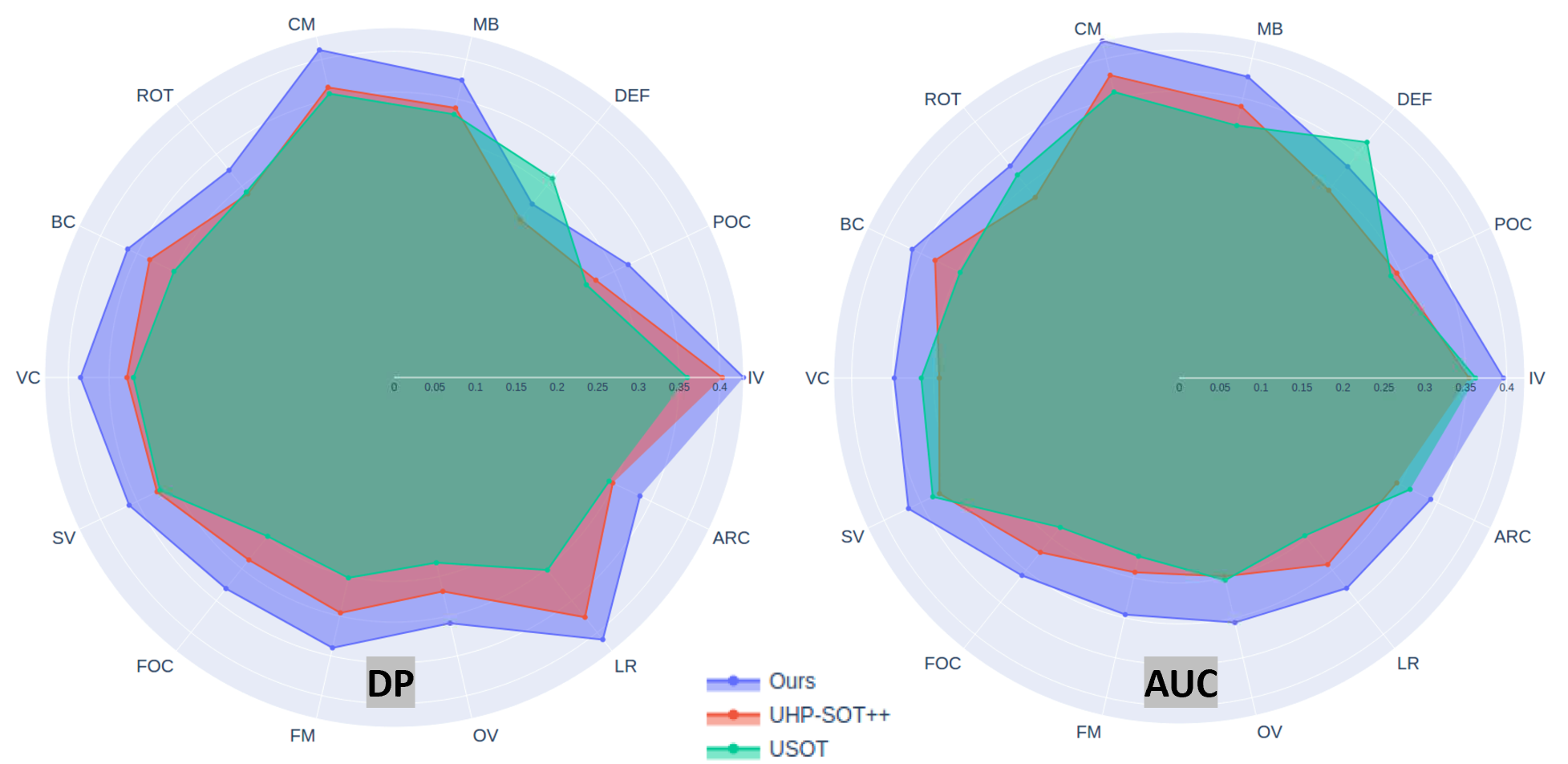}}
\caption{Attribute-based evaluation of GUSOT, UHP-SOT++ and USOT on
LaSOT in terms of DP and AUC, where attributes of interest include the
aspect ratio change (ARC), background clutter (BC), camera motion (CM),
deformation (DEF), fast motion (FM), full occlusion (FOC), illumination
variation (IV), low resolution (LR), motion blur (MB), occlusion (OCC),
out-of-view (OV), partial occlusion (POC), rotation (ROT), scale
variation (SV) and viewpoint change (VC).}\label{fig:ablation2}
\end{figure}

{\bf Performance Evaluation.} We compare tracking accuracy and model
complexity of GUSOT against state-of-the-art supervised and unsupervised
trackers in Table \ref{tab1:comp}. We have the following observations.
First, the improvement over the UHP-SOT++ baseline is around 10\% in DP
and 12\% in AUC. It demonstrates the capability of GUSOT in handling
tracking loss and shape deformation in long video sequences. Second,
GUSOT outperforms all previous DCF-trackers (e.g., supervised ECO and
unsupervised ECO-HC and STRCF) in the first group by a large margin.
Third, GUSOT outperforms two unsupervised deep trackers with
pre-training (e.g. LUDT and USOT) in the second group. Fourth, there is
a significant performance gap between GUSOT and the latest unsupervised
deep tracker ULAST.  Yet, the latter has to be pre-trained by a large
amount of video data with pseudo labels generated from optical flows.
Its large feature backbone, ResNet-50, could be heavy for small devices.
Finally, GUSOT surpasses two supervised deep trackers (i.e., SiamFC and
ECO) in the last group, narrowing the performance gap to the supervised
deep tracker SiamRPN. 

Six examples are selected and shown in Fig.~\ref{fig:vis} for
qualitative comparison.  UHP-SOT++ does not perform well in any of them.
In contrast, GUSOT offers the best performance in all six cases. It
shows the power of the two newly added modules. GUSOT offers flexible
box shapes, yield robust tracking against occlusion and fast motion, and
works well on small objects (e.g., yoyo) and large objects (e.g.,
person). USOT could be distracted by background clutters (see pool,
umbrella, yoyo and person). SiamFC also fails in three right subfigures
(e.g., airplane, person and yoyo). 

We also compare the DP and AUC scores of GUSOT, UHP-SOT++ and USOT
against various tracking attributes in Fig.~\ref{fig:ablation2}. While
USOT performs best in deformation due to its box regression neural
network, GUSOT has leading performance in all other attributes. The
improvement of GUSOT over UHP-SOT++ is more obvious in fast motion,
out-of-view and viewpoint change, which aligns well with the
contributions of the lost object recovery and the color-saliency-based
shape proposal modules. GUSOT can recover the object and adapt the shape
to different viewpoints.

{\bf Ablation Study.} We conduct ablation study on the contributions of
the two proposed modules in Table \ref{tab1:ablation}. Both offer
performance gains in AUC and DP when applied alone.  The shape proposal
contributes more than the motion prosoal in long term tracking.
Furthermore, we incorporate the two modules in three DCF trackers (i.e.
KCF \cite{henriques2014high}, STRCF and UHP-SOT++) and measure their
gains in AUD and DP in Table \ref{tab:dif_baseline}. We see that the two
modules can improve the performance of all three by a significant
amount.  This demonstrates the robustness of the two proposed modules. 

\begin{table}[htbp]
\caption{Ablation study of GUSOT on LaSOT.}\label{tab1:ablation}
\begin{center}
\begin{tabular}{ccccc}\hline
     & baseline & w. motion & w. shape & w. both\\ \hline
AUC (\%) & 32.9 & 35.1 & 36.1 & 36.8 \\
DP (\%) & 32.9 & 34.6 & 35.1 & 36.1 \\ \hline
\end{tabular}
\end{center}
\end{table}

\begin{table}[htbp]
\caption{Performance gain of two new modules on 
different baselines.}\label{tab:dif_baseline}
\begin{center}
\begin{tabular}{cccc} \hline
         & KCF & STRCF & UHP-SOT++ \\ \hline
AUC (\%) & 3.5 & 2.9 & 3.9  \\
DP (\%)  & 3.5 & 2.2 & 3.2 \\ \hline
\end{tabular}
\end{center}
\end{table}

\section{Conclusion and Future Work}\label{sec:conclusion}

A green and unsupervised single-object tracker (GUSOT) for long video
sequences was proposed in this work. As shown by results on LaSOT, the
lost-object-discovery motion proposal and color-saliency-based shape
proposal contributes to unsupervised lightweight trackers significantly.
It offers a promising high-performance tracking solution in mobile and
edge computing platforms. In the future, we would like to introduce
self-supervision to GUSOT to achieve even better tracking performance
while preserving its green and unsupervised characteristics. 


\newpage
\bibliographystyle{IEEEtran}
\bibliography{ref}

\end{document}